\documentclass[10pt,twocolumn,letterpaper]{article}

\usepackage[letterpaper,top=1in,bottom=0.9in,left=0.7in,right=0.7in,
            columnsep=0.25in,headsep=14pt,footskip=30pt]{geometry}

\usepackage[T1]{fontenc}
\usepackage[utf8]{inputenc}
\usepackage{mathptmx}     
\usepackage{textcomp}
\usepackage{microtype}

\usepackage{amsmath,amssymb,amsfonts}
\usepackage{graphicx}
\usepackage{booktabs}
\usepackage{tabularx}
\usepackage{array}
\usepackage{url}
\usepackage{xcolor}

\usepackage[numbers,sort&compress]{natbib}

\usepackage[font=small,labelfont=bf,labelsep=period,skip=4pt]{caption}

\usepackage{titlesec}
\titleformat{\section}{\large\bfseries}{\thesection.}{0.6em}{}
\titleformat{\subsection}{\normalsize\bfseries}{\thesubsection.}{0.6em}{}
\titleformat{\subsubsection}{\normalsize\itshape}{\thesubsubsection.}{0.6em}{}
\titlespacing*{\section}{0pt}{11pt plus 3pt minus 2pt}{5pt}
\titlespacing*{\subsection}{0pt}{8pt plus 2pt minus 2pt}{3pt}

\usepackage{fancyhdr}
\fancypagestyle{firstpage}{%
  \fancyhf{}%
  \fancyfoot[C]{\footnotesize\thepage}%
}

\definecolor{FireBrick}{rgb}{0.698,0.133,0.133}
\definecolor{DarkGreen}{rgb}{0.000,0.392,0.000}
\definecolor{teal}{rgb}{0.000,0.500,0.500}
\definecolor{orange}{rgb}{1.000,0.500,0.000}
\definecolor{nblue}{rgb}{0.000,0.250,0.600}   

\usepackage[
    colorlinks=true,
    citecolor=FireBrick,
    linkcolor=FireBrick,
    urlcolor=nblue,
    linktocpage=true,
    unicode=true,
    pdftitle={Evaluation Principles for MRI-MRA Registration in Trigeminal Neuralgia: An ROI-Centered Neurovascular Benchmark},
    pdfauthor={Xupeng Zhang, Xihang Wang, Michael Xie, Haoyuan Liang, Hau Ern Lien, Oishika Das, James Feghali, Risheng Xu, Peirong Liu}
]{hyperref}

\def\BibTeX{{\rm B\kern-.05em{\sc i\kern-.025em b}\kern-.08em
    T\kern-.1667em\lower.7ex\hbox{E}\kern-.125emX}}

\begin{document}

\sloppy

\twocolumn[{%
\begin{center}
  {\LARGE\bfseries Evaluation Principles for MRI–MRA Registration in Trigeminal Neuralgia: An ROI-Centered Neurovascular Benchmark\par}
  \vspace{14pt}
  {\normalsize
   Xupeng~Zhang$^{1}$,
   Xihang~Wang$^{2}$,
   Michael~Xie$^{2}$,
   Haoyuan~Liang$^{3}$,
   Hau~Ern~Lien$^{3}$,
   Oishika~Das$^{2}$,
   James~Feghali$^{2}$,
   Risheng~Xu$^{2}$, and
   Peirong~Liu$^{1,*}$\par}
  \vspace{8pt}
  {\small
   $^{1}$Department of Electrical and Computer Engineering and Data Science and AI Institute, Johns Hopkins University, Baltimore, MD 21218, USA\\
   $^{2}$Department of Neurosurgery, Johns Hopkins University School of Medicine, Baltimore, MD 21287, USA\\
   $^{3}$Department of Biomedical Engineering, Johns Hopkins University, Baltimore, MD 21287, USA\\[3pt]
   $^{*}$Corresponding author: Peirong~Liu (email: peirong@jhu.edu)\par}
\end{center}
\vspace{10pt}
\begin{center}
\begin{minipage}{0.92\textwidth}
  \small
  \noindent\textbf{Abstract.}\enspace
Preoperative evaluation of trigeminal neuralgia (TN) often requires joint interpretation of structural MRI, which depicts the trigeminal nerve and surrounding cisternal anatomy, and time-of-flight MRA, which highlights vascular structures. Although MRI–MRA fusion is clinically attractive for visualizing neurovascular compression, this task is poorly captured by conventional whole-brain registration evaluation because the clinically relevant target is a small trigeminal ROI, vessel annotations are partial and clinically focused, local TOF-MRA contrast is variable, and field-of-view mismatch can limit deformable alignment. We formulate TN MRI–MRA fusion as an ROI-centered neurovascular registration-evaluation problem and construct a benchmark from 149 patients with clinician-annotated bilateral trigeminal ROIs. Six representative registration pipelines were evaluated using local image-based metrics, segmentation-derived vessel-localization metrics, prediction-volume analysis, and contrast- and FOV-stratified comparisons. Conventional evaluation summaries were often misleading: local image similarity, vessel-background separability, and downstream vessel localization did not co-rank methods; one-sided vessel distances were strongly affected by predicted vessel extent under partial annotations; and local MRA contrast determined when vessel-separability metrics were informative. Deformable refinement provided only a small, FOV-dependent benefit over affine alignment, while reader review showed that locally favorable vessel distances could coexist with globally implausible registrations. These findings indicate that TN MRI–MRA registration should be evaluated as a local, vessel-aware, contrast-sensitive, and FOV-aware visualization task rather than as generic multimodal brain registration. Our code is publicly available at \url{https://github.com/jhuldr/TN-Reg-Benchmark}.
  \par\medskip
  \noindent\textbf{Keywords:}\enspace
Trigeminal neuralgia, MRI-MRA registration, neurovascular compression, vessel segmentation. 
  \par
\end{minipage}
\end{center}
\vspace{18pt}
}]
\thispagestyle{firstpage}

\section{Introduction}

Trigeminal neuralgia (TN) is a debilitating facial pain disorder commonly caused by neurovascular compression (NVC), in which a blood vessel contacts or compresses the trigeminal nerve along its cisternal segment. Preoperative evaluation relies on skull-base imaging that visualizes both the trigeminal nerve and surrounding vasculature. Structural MRI, particularly CISS-type imaging, depicts the trigeminal nerve and adjacent cisternal anatomy, whereas time-of-flight magnetic resonance angiography (TOF-MRA) highlights vascular structures. In current clinical workflows, these modalities are often reviewed side-by-side rather than spatially integrated, requiring clinicians to synthesize complementary anatomical information during interpretation. Accurate MRI-MRA co-registration could therefore support multimodal visualization of nerve-vessel relationships. However, TN MRI-MRA fusion is not a generic whole-brain registration problem: the clinically relevant endpoint is local neurovascular interpretability within a small trigeminal region of interest (ROI), and conventional registration summaries may not reflect whether the fused images are useful for evaluating NVC.

\subsection{MRI–MRA Fusion as a Local Neurovascular Task}

Structural MRI provides high-resolution visualization of the trigeminal nerve and surrounding cisternal anatomy, including vessel-related signal voids near the nerve~\cite{Xu2022MRIRoleNVC}. This supports visual identification of NVC and segmentation-based assessment of NVC morphology and pain-outcome modeling~\cite{HalbertElliott2025Segmentation,Wang2026NVCQuantification}. However, structural MRI alone has limited nerve–vessel contrast, incomplete vascular conspicuity, and limited vessel-type information. TOF-MRA complementarily emphasizes vascular structures and can help distinguish vessels from surrounding soft tissue. Because TOF-MRA preferentially depicts arterial flow, TOF-visible vessels may also provide indirect information about vascular identity, although TOF visibility alone is not definitive.

The goal of TN's MRI–MRA fusion is therefore not simply whole-brain alignment, but spatial integration of vascular MRA with structural MRI near the trigeminal nerve. A registration that appears acceptable globally may still fail within the trigeminal ROI, while a favorable local vessel metric may be misleading if global alignment is implausible. This motivates ROI-centered evaluation focused on local neurovascular interpretability rather than whole-volume similarity alone.

\subsection{Clinical Use Case}

We target preoperative evaluation for microvascular decompression (MVD), which treats TN by separating contacting vessel from the trigeminal nerve. Accurate visualization of the nerve and candidate offending vessels is clinically important as NVC morphology and vessel type are associated with postoperative pain outcomes, including contact location, contact severity or surface area, and arterial versus venous compression~\cite{Mistry2016REZ,Loayza2023OutcomeNVC,Nair2023ArteryVein,Wang2026NVCQuantification}. Spatially accurate MRI–MRA fusion may therefore support morphologic assessment of nerve–vessel contact within the trigeminal ROI. The immediate technical goal is not to determine vessel type or predict outcome, but to produce and evaluate anatomically plausible multimodal visualization.

\subsection{Why Conventional Registration Evaluation Is Insufficient}

Although MRI and TOF-MRA play complementary roles in TN evaluation, MRI–MRA co-registration remains underexplored in TN-specific applications. Prior TN studies often review modalities separately or demonstrate fusion qualitatively, while most registration benchmarks target generic brain alignment rather than clinician-defined trigeminal ROIs. TN MRI–MRA fusion challenges conventional evaluation in four ways: the endpoint is a small local ROI rather than the full brain; vessel annotations are partial and clinically focused, making one-sided distances sensitive to predicted vessel volume; local TOF-MRA conspicuity varies, limiting vessel-separability metrics in low-contrast ROIs; and MRI–MRA FOV mismatch can limit deformable refinement, increasing the importance of robust affine initialization and quality control.

\subsection{Related Work}

\textbf{Registration methodology.} Classical deformable registration relies on similarity-driven optimization with explicit transformation models. ANTs~\cite{Avants2008SyN,Avants2011ANTs,Tustison2021ANTsX} remains a widely used standard for affine and diffeomorphic registration. FireANTs~\cite{Jena2024FireANTs} accelerates diffeomorphic matching on GPU, and ConvexAdam~\cite{Siebert2025ConvexAdam} combines discrete and continuous optimization but assumes approximate affine alignment. Learning-based methods such as EasyReg~\cite{Iglesias2023EasyReg} and SynthMorph~\cite{Hoffmann2022SynthMorph} aim to improve robustness and inference speed. Downstream vessel extraction can provide task-specific localization readouts; here, we use VesselFM~\cite{Wittmann2025VesselFM} as the segmentation component. These methods are typically evaluated on whole-brain alignment tasks with adequate anatomical overlap, and their reliability for local TN visualization under FOV mismatch, variable MRA conspicuity, and partial vessel annotations remains unclear.

\textbf{Clinical imaging for TN and NVC.} Preoperative MRI and MRA are central to TN evaluation, but standardized objective NVC grading remains challenging, and NVC identification alone is not a complete surrogate for symptoms or recurrence~\cite{Lee2014AbsentNVC}. NVC severity, compression location, nerve atrophy, and vessel type have been associated with clinical presentation or MVD outcomes~\cite{Hughes2019NVCSeverity,Worm2025FiveYear,Loayza2023OutcomeNVC,Antonini2014SymptomaticNVC,Mistry2016REZ,Leal2014NerveAtrophy,Liu2017NerveAtrophy,Nair2023ArteryVein}. Prior studies have demonstrated fusion of high-resolution structural MRI with TOF-MRA for posterior-fossa neurovascular visualization and surgical planning~\cite{Miller2008PreoperativeVisualization,Docampo2015NeurovascularStudy3T,Granata2013VirtualMRI,Dolati2015PreoperativeSegmentation,Yao2018VirtualRealityMVD,Gamaleldin2020FusedTOFCISS,Pham2021FusedSpaceTOFMRA,Hastreiter2022DataFusion3D,Huang2024MRVE}, including ultra-high-field multimodal MRI in secondary TN~\cite{ArrighiAllisan2020SecondaryTN7T}. These studies motivate multimodal visualization, but do not define how registration should be evaluated when the endpoint is local vessel interpretability in a clinician-defined trigeminal ROI.

\subsection{Contributions}

This work makes four contributions. First, we formulate TN MRI-MRA fusion as an ROI-centered neurovascular registration-evaluation problem rather than a generic whole-brain alignment task. Second, we assemble a clinical cohort of 149 TN patients with clinician-annotated bilateral trigeminal ROIs and link whole-brain registration outputs to local vessel-focused evaluation. Third, we evaluate six representative registration pipelines using complementary image-based metrics and segmentation-derived downstream localization metrics, explicitly separating local vessel separability, downstream vessel extraction, and global registration plausibility. Fourth, through predicted-volume, contrast-stratified, FOV-stratified, and reader-based analyses, we identify practical confounds missed by conventional evaluation and define evaluation principles for future TN-specific MRI-MRA registration.

\section{Materials and Methods}
\begin{figure*}[t]
    \centering
    \includegraphics[width=0.9\textwidth]{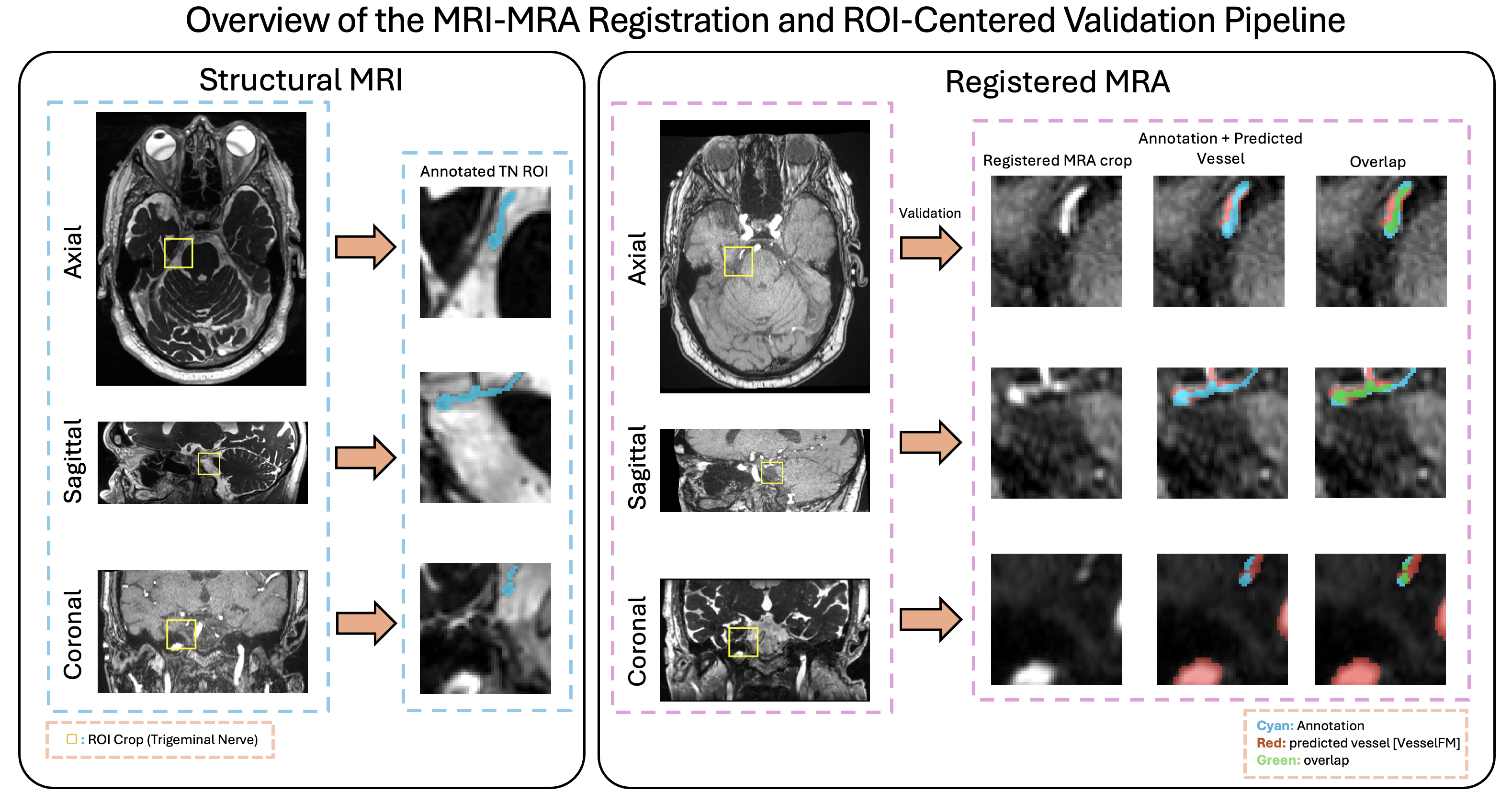}
    \caption{\textbf{Overview of the MRI–MRA registration and ROI-centered evaluation pipeline.} MRI provides the anatomical reference and clinician-annotated trigeminal ROI; MRA is registered into MRI space. Local ROIs are extracted from the warped MRA for segmentation-based vessel validation. Cyan: clinician-labeled vessel segment; red: predicted vessel; green: overlap.}
    \label{fig:pipeline}
\end{figure*}

\subsection{Clinical Cohort}

We retrospectively studied a single-institution cohort of 149 TN patients (298 bilateral trigeminal ROIs) who underwent diagnostic structural MRI and TOF-MRA between October 2019 and December 2022 and subsequently underwent MVD. Inclusion required paired structural MRI and TOF-MRA from the same preoperative imaging encounter and clinician-provided trigeminal annotations. The final cohort had a mean age of $56.6 \pm 14.0$ years (85 female / 64 male; 88 right- / 61 left-sided presentation; 7 patients [4.7\%] with multiple sclerosis). The cohort flow from 159 initial paired scans to 149 analyzable patients, and the per-analysis subset counts, are summarized in Table~\ref{tab:cohort_flow}. The study was approved by the Johns Hopkins Medicine Institutional Review Board (IRB00338945); informed consent was waived given the retrospective design.

\begin{table}[t]
\centering
\caption{Cohort flow from initial paired scans to final analysis subsets.}
\label{tab:cohort_flow}
\small
\setlength{\tabcolsep}{3pt}
\renewcommand{\arraystretch}{1.05}
\begin{tabular}{@{}lccl@{}}
\toprule
\textbf{Analysis step} & \textbf{Patients} & \textbf{ROIs} & \textbf{Excluded} \\
\midrule
Initial paired scans         & 159 & -  & - \\
Preprocessed pairs           & 153 & -  & 6 (preprocessing) \\
Final analysis cohort        & 149 & 298 & 4 (MRA unpaired) \\
Registration outputs         & 149 & 298 & 0 \\
Contrast-stratified          & 149 & 286 & 12 (empty SyN ROI) \\
FOV paired     & 149 & 223 & 75 (empty/out-of-FOV) \\
\bottomrule
\end{tabular}
\end{table}

\subsection{Imaging Data and ROI Annotations}

Structural MRI used a 3D CISS-type sequence ($\sim$0.6mm isotropic), and TOF-MRA used a 3D multi-slab sequence ($\sim$0.26mm in-plane, $\sim$0.5mm slice). The cohort was predominantly Siemens 3T: 98.0\% Siemens; structural MRI included 138/149 scans at 3T and 11/149 at 1.5T; TOF-MRA included 137/149 scans at 3T and 12/149 at 1.5T.

Local annotations were available within $48 \times 48 \times 48$ ROI crops centered on the cisternal trigeminal nerve, sampled on a $0.47$mm isotropic grid. Each ROI contained per-voxel labels for the trigeminal nerve and vessel structures, and all patients had both ipsilateral and contralateral ROIs annotated. Annotations were drawn on structural MRI alone, independent of MRA and registration outputs, using a two-pass workflow consisting of initial segmentation followed by independent review. Vessel labels marked only the segment considered clinically relevant to the trigeminal nerve, not an exhaustive vascular tree. This partial-annotation property is central to metric interpretation and is analyzed quantitatively in Section~\ref{sec:volume-coupling}.

\subsection{Preprocessing and Registration Methods}
All images were converted from DICOM to NIfTI and resampled to RAS orientation. Brain masks were estimated to suppress non-brain background and support automatic cropping. MRI served as the fixed image and MRA as the moving image; each method resampled MRA into MRI space.

We evaluated six registration methods: ANTs Affine~\cite{Avants2011ANTs,Tustison2021ANTsX}, ANTs SyN~\cite{Avants2008SyN}, ConvexAdam~\cite{Siebert2025ConvexAdam}, FireANTs~\cite{Jena2024FireANTs}, EasyReg~\cite{Iglesias2023EasyReg}, and SynthMorph~\cite{Hoffmann2022SynthMorph}. In figures, they are abbreviated as Affine, SyN, CvxAd, Fire, Easy, and Synth. Table~\ref{tab:methods} summarizes their categories and initialization protocols.

\textbf{Pre-alignment protocol.} ANTs SyN, EasyReg, and SynthMorph include affine alignment within their workflows. ConvexAdam and FireANTs were supplied with the ANTs Affine output as a unified external pre-alignment, because they were not treated here as standalone cross-modality affine pipelines for MRI–MRA. Per-method configuration details, including similarity metrics, multi-resolution schedules, descriptors, normalization, software versions, and hardware, are provided in the released code repository.

\begin{table*}[t]
\centering
\caption{Summary of the registration methods evaluated in this study.}
\label{tab:methods}
\small
\begin{tabular}{lllll}
\toprule
\textbf{Method} & \textbf{Category} & \textbf{Initialization} & \textbf{GPU} & \textbf{Notes} \\
\midrule
ANTs Affine & Affine               & ANTs affine                        & No  & Global affine baseline. \\
ANTs SyN    & Classical deformable & Internal (ANTs affine + SyN)       & No  & Diffeomorphic refinement after affine. \\
ConvexAdam  & Optimization-based   & External ANTs affine               & Yes & Requires affine pre-alignment. \\
FireANTs    & Optimization-based   & External (ANTs affine + FireANTs)  & Yes & GPU-accelerated; not standalone affine. \\
EasyReg     & Learning-based       & Internal                           & No  & Segmentation-guided contrast-agnostic. \\
SynthMorph  & Learning-based       & Internal                           & No  & Synthetic-training contrast-invariant. \\
\bottomrule
\end{tabular}
\end{table*}

\subsection{ROI-Centered Segmentation-Based Evaluation}
After whole-brain registration, the MRA was linked to the annotated trigeminal ROI using physical-space centroids and a fixed local sampling grid~(Fig.~\ref{fig:pipeline}). Metrics were computed within the matched $48^3$ ROI at $0.47$ mm isotropic resolution.

For segmentation-based vessel evaluation, we applied VesselFM~\cite{Wittmann2025VesselFM} (v1.0, \texttt{dyn\_unet\_base}, no retraining) to each warped MRA volume in MRI space, then cropped the predicted whole-brain vessel mask to the matched trigeminal ROI. The same inference configuration was used for all warped MRA outputs, so cross-method differences reflect changes induced by registration rather than changes in the segmentation pipeline. Vessel-proximity metrics required at least one predicted vessel voxel in the ROI; observations with empty vessel predictions were excluded only from those metrics. Because predicted vessel extent varied substantially across methods, we tracked predicted vessel volume as an auxiliary quantity and analyzed its effect on distance metrics in Section~\ref{sec:volume-coupling}. Full preprocessing, inference, binarization, and postprocessing details are provided in the released code repository.

\subsection{Stratified Analyses}
\label{sec:strat-methods}
\textbf{Contrast stratification.} For each ROI, we computed a local vessel-to-background intensity contrast score on warped MRA,
\begin{equation}
\mathrm{contrast} = \bar{I}_{\mathrm{MRA}}(V_{\mathrm{ann}}) / \bar{I}_{\mathrm{MRA}}(B),
\label{eq:contrast}
\end{equation}
$V_{\mathrm{ann}}$ and $B$ denote the clinician-labeled vessel voxels and the background voxels (labeled neither vessel nor trigeminal nerve), respectively, within ROI. The contrast score was computed on the ANTs SyN warped MRA and used as a canonical per-ROI reference. ROIs were stratified into Low ($\mathrm{contrast} \le 1.1$, $N=181$, 63\%), Mid ($1.1 < \mathrm{contrast} \le 1.3$, $N=66$, 23\%), and High ($\mathrm{contrast} > 1.3$, $N=39$, 14\%) tiers; 12 of 298 ROIs were excluded as ANTs SyN deformable warp placed the TN-centered sampling grid outside the warped MRA coverage.

\textbf{FOV stratification.} MRI-MRA FOV compatibility was characterized using a per-case worst-axis coverage ratio,
\begin{equation}
r_{\min} = \min_{a \in \{x, y, z\}} \frac{\mathrm{FOV}_a^{\mathrm{MRI}}}{\mathrm{FOV}_a^{\mathrm{MRA}}},\quad \mathrm{FOV}_a = \mathrm{shape}_a \times \mathrm{spacing}_a .
\label{eq:fov}
\end{equation}
Cases with $r_{\min} \le 0.76$ (the cohort's 20th percentile) were classified as Bad FOV, and the remainder as Good FOV. The Affine-versus-SyN paired analysis included 177 Good-FOV and 46 Bad-FOV ROIs, after excluding 75 ROIs with empty vessel predictions or out-of-coverage warped regions. For descriptive cross-tabulation, we defined image-level high-AUC as $\mathrm{AUC}_{\mathrm{SyN}} > 0.60$ and downstream miss as $d_{\mathrm{ann}\to\mathrm{Pred}} > 2$mm. Volume-distance coupling was assessed by Spearman correlation, and paired Affine-versus-SyN comparisons used the Wilcoxon signed-rank test. Confidence intervals were estimated by patient-level bootstrap resampling with 1000 resamples to account for bilateral ROIs. 

\subsection{Blinded Reader Study}
\label{sec:reader-study}
To relate ROI-centered metrics to clinical interpretability and to probe whether low local TOF-MRA contrast may reflect venous-offender biology, we conducted a blinded reader study on 100 stratified trigeminal ROIs. The sample included 50 Low-contrast ROIs, 20 Mid-contrast ROIs with downstream vessel-localization misses, 20 High-contrast hit controls, and 10 harder Mid/High hit controls. ROIs were randomized as R001–R100, and the manifest linking review ID to patient identity, side, and sampling stratum was withheld during review.

Two attending neurosurgeons assessed each ROI jointly in consensus using the structural MRI, original MRA, ANTs-SyN warped MRA, and the single-side ROI label. For each ROI, they recorded clinical evaluability, compressive vessel type, and confidence. An ROI was considered evaluable if the ANTs-SyN warped MRA was anatomically consistent with the structural MRI and usable for assessing the neurovascular relationship; globally implausible overlays were rated non-evaluable. For evaluable ROIs, vessel type was recorded as artery, vein, mixed, or no definite vessel. Because TOF-MRA preferentially depicts arterial flow and may not show slow venous flow, vessel type was assigned by tracing candidate vessels across adjacent slices and assessing continuity with named posterior-fossa vessels, rather than by TOF signal alone. Arterial candidates were assessed for continuity with the superior cerebellar or anterior inferior cerebellar artery, whereas vessels lacking arterial continuity and following a typical petrosal venous course were classified as venous. SynthMorph outputs were additionally tracked as a method-specific registration-quality comparison in a separate field, so SynthMorph-specific failures did not affect the primary ANTs-SyN evaluability labels.

\section{Experimental Results}

\subsection{Evaluation Metrics}
\label{sec:eval-metrics}
We evaluated registration outputs within the clinician-defined $48^3$ trigeminal ROI using complementary metrics that capture different aspects of the TN MRI–MRA fusion task. Let $V_{\mathrm{ann}}$ denote the clinician-labeled vessel voxels within the ROI, $V_{\mathrm{Pred}}$ the segmentation-predicted vessel voxels, and $B$ the background voxels labeled as neither vessel nor trigeminal nerve. The annotation $V_{\mathrm{ann}}$ is a clinically focused partial vessel segment rather than an exhaustive vascular tree, which is central to the interpretation of the distance metrics below. Let $I_{\mathrm{MRA}}(v)$ denote the warped MRA intensity at voxel $v$.

\textbf{Image-based metrics.} \emph{Vessel AUC} treats the registered MRA intensity as a score for separating clinician-labeled vessel voxels from background voxels within the ROI:
\begin{equation}
\mathrm{AUC} = P\big(I_{\mathrm{MRA}}(v_{+}) > I_{\mathrm{MRA}}(v_{-})\big),\quad v_{+} \in V_{\mathrm{ann}},\; v_{-} \in B .
\label{eq:auc}
\end{equation}
A value of $0.5$ corresponds to chance-level separability. \emph{ROI NMI} measures local multimodal intensity correspondence between the fixed MRI and warped MRA:
\begin{equation}
\mathrm{NMI}(X, Y) = (H(X) + H(Y)) / {H(X, Y)},
\label{eq:nmi}
\end{equation}
where $X$ and $Y$ denote the intensity distributions of the fixed MRI and warped MRA within the ROI, respectively.

\textbf{Vessel-proximity metrics.} To evaluate downstream vessel localization, we computed mean nearest-neighbor distances between the annotation and post-registration vessel prediction:
\begin{align}
d_{\mathrm{ann}\to\mathrm{Pred}} &= {1} / {|V_{\mathrm{ann}}|} \sum_{x \in V_{\mathrm{ann}}} \min_{y \in V_{\mathrm{Pred}}} \|x - y\|_2,
\label{eq:gt2pred} \\
d_{\mathrm{Pred}\to\mathrm{ann}} &= {1} / {|V_{\mathrm{Pred}}|} \sum_{y \in V_{\mathrm{Pred}}} \min_{x \in V_{\mathrm{ann}}} \|x - y\|_2,
\label{eq:pred2gt} \\
d_{\mathrm{sym}} &= \tfrac{1}{2}\,(d_{\mathrm{ann}\to\mathrm{Pred}} + d_{\mathrm{Pred}\to\mathrm{ann}}).
\label{eq:sym}
\end{align}
Because the annotation is partial, $d_{\mathrm{ann}\to\mathrm{Pred}}$ measures whether the clinically annotated segment is covered by the predicted vessel tree, whereas $d_{\mathrm{Pred}\to\mathrm{ann}}$ penalizes predicted vessels that extend beyond the annotated segment. We therefore report both directional distances, their symmetric average, and the predicted vessel volume $|V_{\mathrm{Pred}}|$.

\subsection{Cohort and Analysis Subsets}
\label{sec:cohort-summary}

The final cohort included 149 TN patients and 298 bilateral trigeminal ROIs. All registration methods produced whole-brain warped MRA outputs for all patients. Image-based metrics were computed on all 298 ROIs. Contrast-stratified analysis used 286 ROIs with valid ANTs SyN-based contrast measurements; 12 ROIs were excluded because the SyN-warped sampling grid fell outside the warped MRA coverage. Vessel-proximity metrics were computed on per-method subsets with non-empty post-registration vessel predictions, ranging from 243 to 277 ROIs depending on the registration method. The FOV-stratified Affine-versus-SyN paired analysis included 223 ROIs, consisting of 177 Good-FOV and 46 Bad-FOV ROIs, after excluding cases with empty predicted vessel masks or out-of-coverage warped regions.

\subsection{Conventional Metrics Reveal Discordant Rankings}
\label{sec:overall-comparison}

\begin{figure}[t]
    \centering
    \includegraphics[width=\columnwidth]{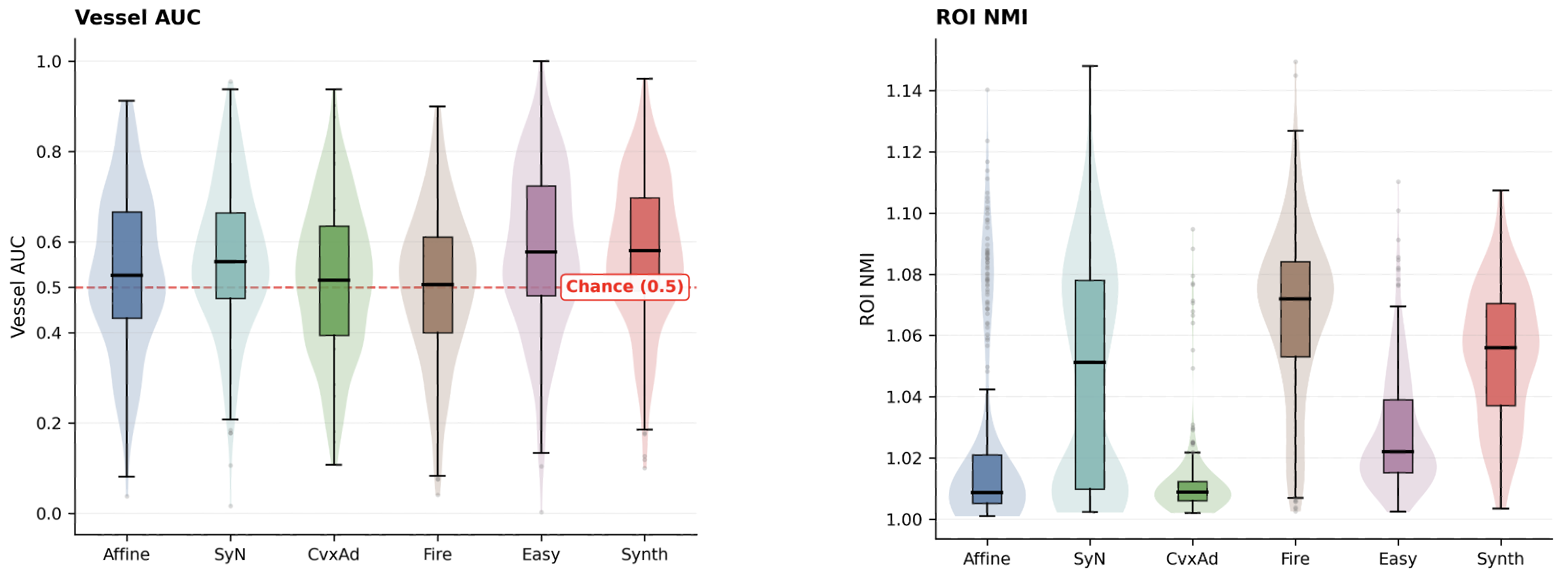}
    \caption{\textbf{Local image-based metrics across registration methods.} (A) Vessel AUC; dashed line indicates chance level ($0.5$). (B) ROI NMI. The two metrics do not co-rank methods, indicating that local vessel separability and local multimodal intensity correspondence capture different properties of the registered images. Numerical values are reported in Table~\ref{tab:overall_metrics}.}
    \label{fig:overall_intensity}
\end{figure}

We first compared the six registration methods using local image-based and downstream vessel-proximity summaries. For vessel-background intensity discrimination (Fig.~\ref{fig:overall_intensity}A), EasyReg and SynthMorph achieved the highest median Vessel AUC values (0.578 and 0.581), then ANTs SyN (0.557), while ANTs Affine, ConvexAdam, and FireANTs were closer to chance. ROI NMI ranked methods differently (Fig.~\ref{fig:overall_intensity}B): FireANTs was highest (1.072), followed by SynthMorph (1.056) and ANTs SyN (1.051), with ANTs Affine and ConvexAdam lowest. Thus, local vessel separability and local multimodal intensity correspondence captured different properties of the registered images rather than a single shared notion of registration quality.

\begin{figure}[t]
    \centering
    \includegraphics[width=\columnwidth]{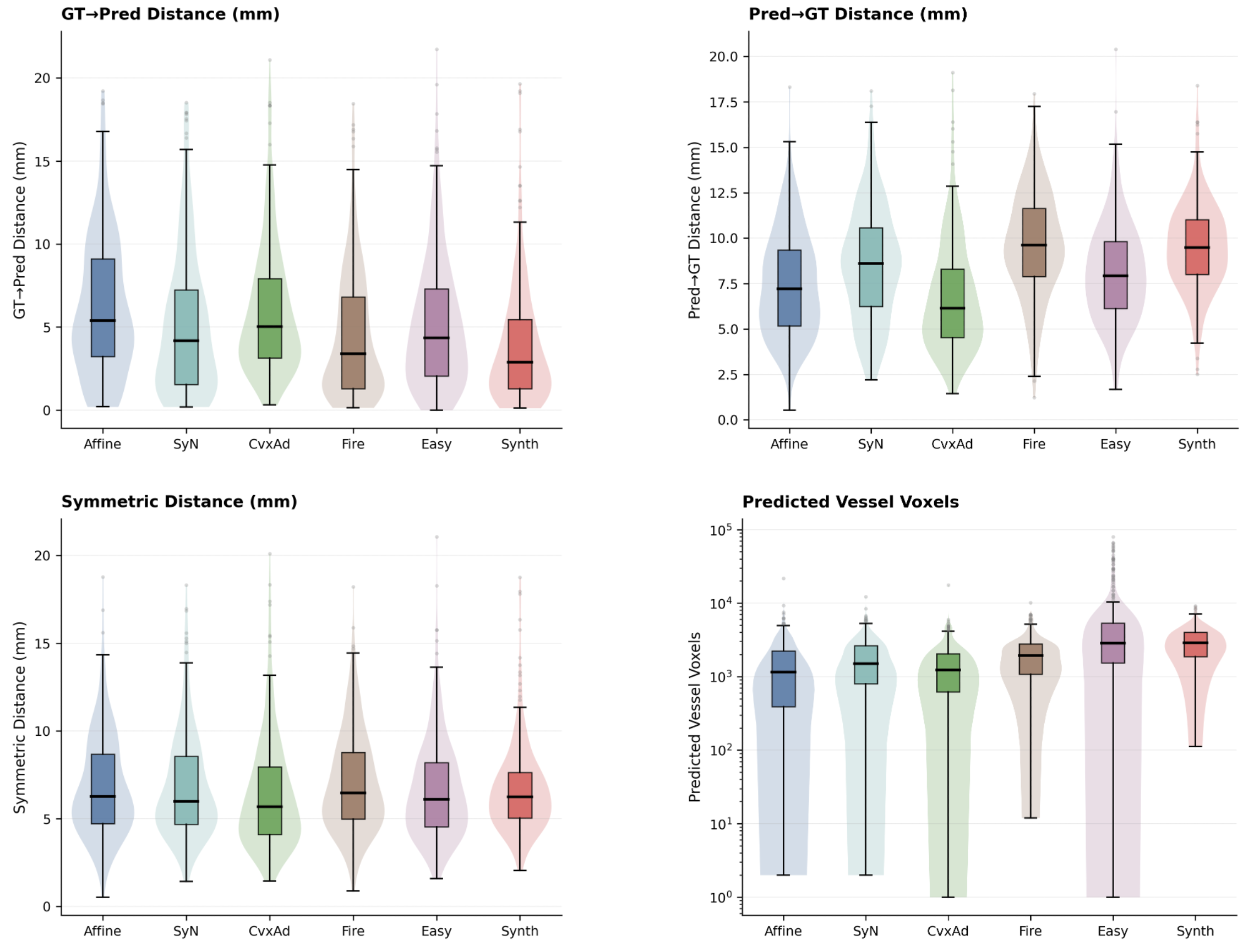}
    \caption{\textbf{Vessel-proximity and prediction-volume metrics.} (A) $d_{\mathrm{ann}\to\mathrm{Pred}}$. (B) $d_{\mathrm{Pred}\to\mathrm{ann}}$. (C) $d_{\mathrm{sym}}$. (D) Predicted vessel voxels $|V_{\mathrm{Pred}}|$ on a log scale. Directional distances rank methods differently, and these differences are coupled to predicted vessel volume. Numerical values are reported in Table~\ref{tab:overall_metrics}.}
    \label{fig:overall_geometry}
\end{figure}

Downstream vessel-proximity metrics showed an additional discordance (Fig.~\ref{fig:overall_geometry}). Under $d_{\mathrm{ann}\to\mathrm{Pred}}$, SynthMorph and FireANTs appeared strongest, whereas under $d_{\mathrm{Pred}\to\mathrm{ann}}$, the ordering reversed and these methods produced the largest distances back to the annotation. This asymmetry coincided with large differences in predicted vessel volume, with EasyReg and SynthMorph producing substantially larger predicted vessel masks than others. Symmetric distance $d_{\mathrm{sym}}$ reduced but did not eliminate this directional dependence. Numerical values are reported in Table~\ref{tab:overall_metrics}. These results show that image similarity, vessel separability, and one-sided vessel localization are not interchangeable measures of TN MRI-MRA registration quality.

\begin{table*}[!t]
\centering
\caption{Local image-based and vessel-proximity metrics across registration methods. Image-based metrics are computed on 298 ROIs. Vessel-proximity metrics are computed on per-method subsets with non-empty post-registration vessel predictions ($N$ in the last column). Distances are in mm; volumes are in voxels. Values are median [Q1, Q3]. See Figs.~\ref{fig:overall_intensity} and \ref{fig:overall_geometry} for distributions.}
\label{tab:overall_metrics}
\small
\setlength{\tabcolsep}{3pt}
\begin{tabular}{lccccccc}
\toprule
& \multicolumn{2}{c}{\textbf{Image-based}} & \multicolumn{5}{c}{\textbf{Vessel-proximity (segmentation-derived)}} \\
\cmidrule(lr){2-3} \cmidrule(lr){4-8}
\textbf{Method} & \textbf{Vessel AUC} & \textbf{ROI NMI} & $d_{\mathrm{ann}\to\mathrm{Pred}}$ \textbf{(mm)} & $d_{\mathrm{Pred}\to\mathrm{ann}}$ \textbf{(mm)} & $d_{\mathrm{sym}}$ \textbf{(mm)} & $|V_{\mathrm{Pred}}|$ \textbf{(voxels)} & $N$ \\
\midrule
ANTs Affine & 0.526 [0.432, 0.667] & 1.009 [1.005, 1.021] & 5.38 [3.23, 9.12] & 7.22 [5.17, 9.35]  & 6.26 [4.71, 8.67] & 1148 [388, 2227]  & 243 \\
ANTs SyN    & 0.557 [0.476, 0.665] & 1.051 [1.010, 1.078] & 4.18 [1.54, 7.24] & 8.60 [6.25, 10.58] & 5.97 [4.67, 8.56] & 1493 [803, 2643]  & 250 \\
ConvexAdam  & 0.522 [0.396, 0.640] & 1.009 [1.006, 1.012] & 4.90 [3.10, 7.86] & 6.16 [4.55, 8.46]  & 5.67 [4.09, 7.84] & 1218 [624, 2039]  & 246 \\
FireANTs    & 0.506 [0.399, 0.611] & 1.072 [1.053, 1.084] & 3.40 [1.29, 6.81] & 9.61 [7.89, 11.64] & 6.46 [4.98, 8.77] & 1928 [1082, 2774] & 277 \\
EasyReg     & 0.578 [0.481, 0.724] & 1.022 [1.015, 1.039] & 4.36 [2.06, 7.30] & 7.93 [6.12, 9.81]  & 6.10 [4.53, 8.19] & 2849 [1529, 5318] & 267 \\
SynthMorph  & 0.581 [0.493, 0.698] & 1.056 [1.037, 1.071] & 2.88 [1.30, 5.45] & 9.49 [8.00, 11.02] & 6.23 [5.03, 7.62] & 2876 [1868, 4000] & 277 \\
\bottomrule
\end{tabular}
\end{table*}

\begin{figure*}[t]
    \centering
    \includegraphics[width=1\textwidth]{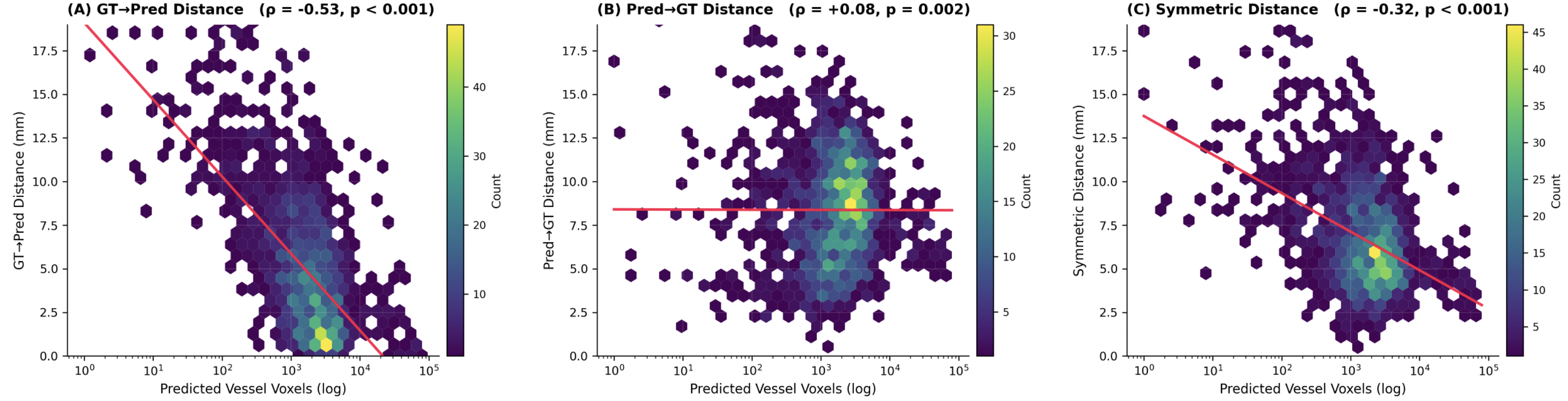}
    \caption{\textbf{Coupling between predicted vessel volume and directional vessel-distance metrics.} Each panel shows the 2D density of per-(method, ROI) observations; red line is a linear regression on log-transformed volume, and $\rho$ is the Spearman correlation. Annotation-to-prediction distance is strongly volume-dependent, whereas prediction-to-annotation distance is not.}
    \label{fig:predvox_bias}
\end{figure*}

\begin{figure*}[!p]
    \centering
    \includegraphics[width=0.76\textwidth]{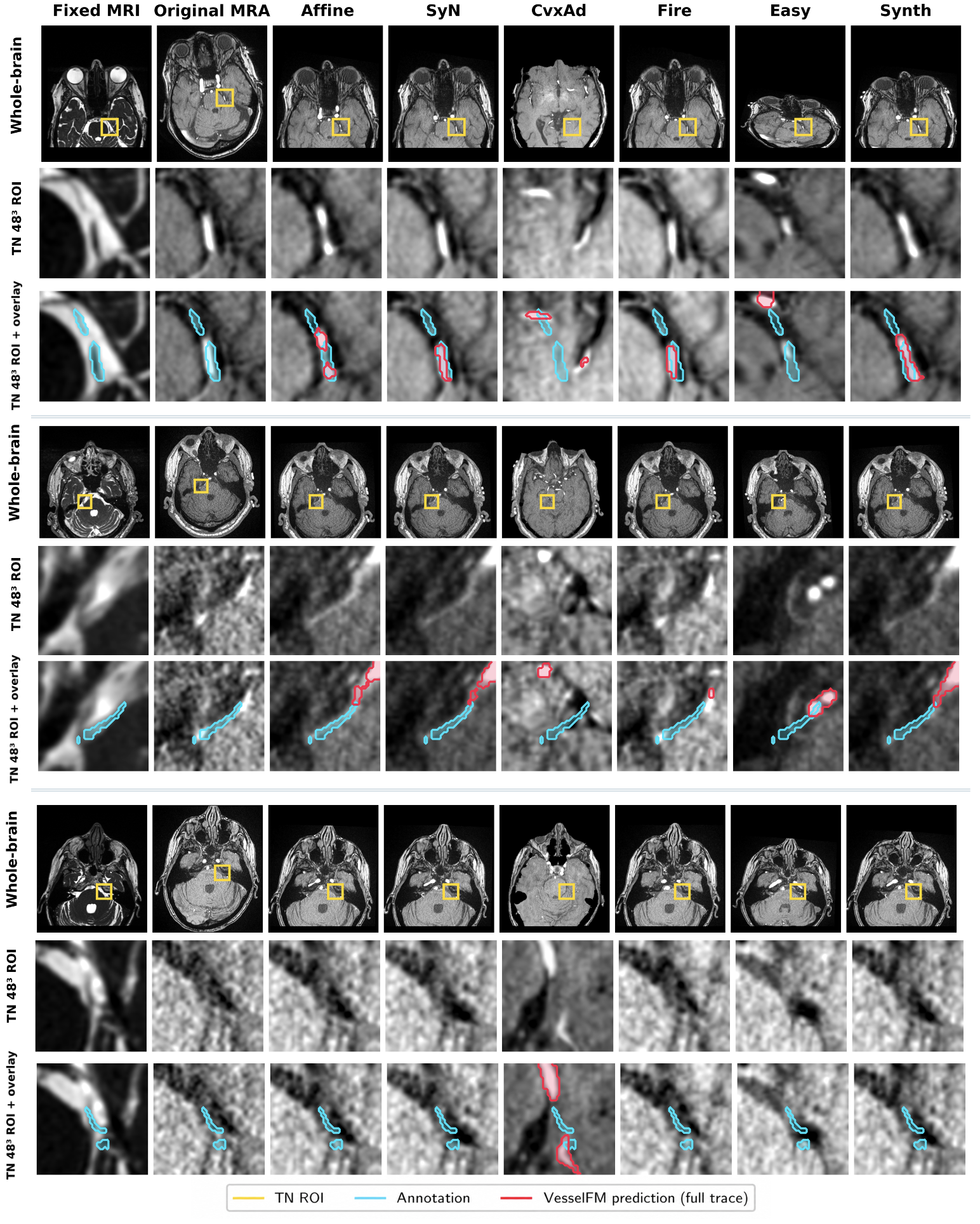}

    \caption{\textbf{Qualitative operating regimes for TN MRI-MRA registration across all six methods.} Each panel (A-C) shows the fixed MRI, original MRA, and warped MRA outputs from ANTs Affine, ANTs SyN, ConvexAdam, FireANTs, EasyReg, and SynthMorph. Rows within each panel show the whole-brain slice (yellow box marks the trigeminal ROI), the matched $48 \times 48 \times 48$ trigeminal ROI intensity, and the ROI overlay (cyan: clinician annotation; red: post-registration vessel prediction).
    \textbf{(A) High-contrast, Good-FOV success case} (Representative Case A, ipsilateral ROI): under ANTs SyN, $\mathrm{AUC}_{\mathrm{SyN}} = 0.88$ and $d_{\mathrm{ann}\to\mathrm{Pred}} = 0.64$mm. Most methods localize the annotated vessel segment, illustrating the favorable operating regime in which local MRA conspicuity and downstream vessel localization agree.
    \textbf{(B) Mid-tier image-level success with downstream vessel-localization miss} (Representative Case B, ipsilateral ROI): despite Good-FOV coverage and measurable image-level separability under ANTs SyN ($\mathrm{AUC}_{\mathrm{SyN}} = 0.70$), segmentation-based predictions are non-empty across methods but spatially displaced from the annotated segment ($d_{\mathrm{ann}\to\mathrm{Pred}} = 6.53$mm under SyN), illustrating the gap between image-level vessel separability and model-level vessel extraction.
    \textbf{(C) Low-contrast image-level failure} (Representative Case C, contralateral ROI): although the whole-brain registrations place the trigeminal ROI in comparable anatomical locations, local vessel-background separability is near chance under ANTs SyN ($\mathrm{AUC}_{\mathrm{SyN}} = 0.50$, contrast $= 0.89$) and the predicted vessel mask does not localize the annotated segment ($d_{\mathrm{ann}\to\mathrm{Pred}} = 6.89$mm).
    All distances are computed in the full 3D ROI, not only on the displayed axial slice. Together, these examples illustrate that whole-brain visual alignment, image-level vessel conspicuity, and downstream model-based vessel localization are related but non-interchangeable evaluation axes, and that this dissociation is not specific to any single registration method.}
    \label{fig:case_example}
\end{figure*}

\subsection{Predicted Volume Confounds One-Sided Distance Metrics}
\label{sec:volume-coupling}
To test the volume-confound suggested by the overall metrics, we examined predicted vessel volume against directional distance metrics across all method-ROI observations (Fig.~\ref{fig:predvox_bias}). $d_{\mathrm{ann}\to\mathrm{Pred}}$ showed a strong negative association with predicted volume (Spearman $\rho = -0.53$, 95\% CI: $[-0.59, -0.47]$), consistent with the intuition that larger predicted vessel trees are more likely to cover the annotated segment. $d_{\mathrm{Pred}\to\mathrm{ann}}$ showed only a negligible association ($\rho = +0.08$, 95\% CI: $[+0.01, +0.15]$), demonstrating that enlarging the prediction does not, on average, bring predicted voxels closer to the annotated vessel. $d_{\mathrm{sym}}$ showed an intermediate association ($\rho = -0.32$, 95\% CI: $[-0.38, -0.25]$). These results indicate that, under partial clinical annotations and cross-method differences in prediction extent, one-sided vessel-proximity metrics can systematically favor methods that produce more voluminous vessel trees, independent of local geometric accuracy. Symmetric distance is less directionally biased than either one-sided distance, but it remains a segmentation-derived downstream localization metric rather than an independent registration-error measure. We therefore report it jointly with the two one-sided distances and prediction-volume statistics.

\subsection{Visibility Does Not Guarantee Downstream Localization}
\label{sec:operating-regimes}

Fig.~\ref{fig:case_example} illustrates three qualitative operating regimes that explain why TN MRI–MRA fusion requires separate evaluation axes. In a high-contrast, Good-FOV case (Fig.~\ref{fig:case_example}A), local vessel signal is conspicuous and segmentation-based prediction localizes the clinician-annotated vessel segment, representing the favorable regime in which image-level visibility and downstream localization agree. In a Mid-tier case (Fig.~\ref{fig:case_example}B), local vessel-background separability remains measurable under ANTs SyN, yet the segmentation-based prediction is non-empty but spatially displaced from the annotation. This demonstrates that image-level vessel visibility does not necessarily imply successful model-based vessel localization. In a Low-contrast case with Good-FOV coverage (Fig.~\ref{fig:case_example}C), vessel-background separability is near chance and downstream localization fails. These examples motivate evaluating TN MRI-MRA registration along separate whole-brain, image-level, and downstream localization axes.

\subsection{Local Contrast Determines When Vessel AUC Is Informative}
\label{sec:contrast-strat}
\begin{figure}[t]
    \centering
    \includegraphics[width=\columnwidth]{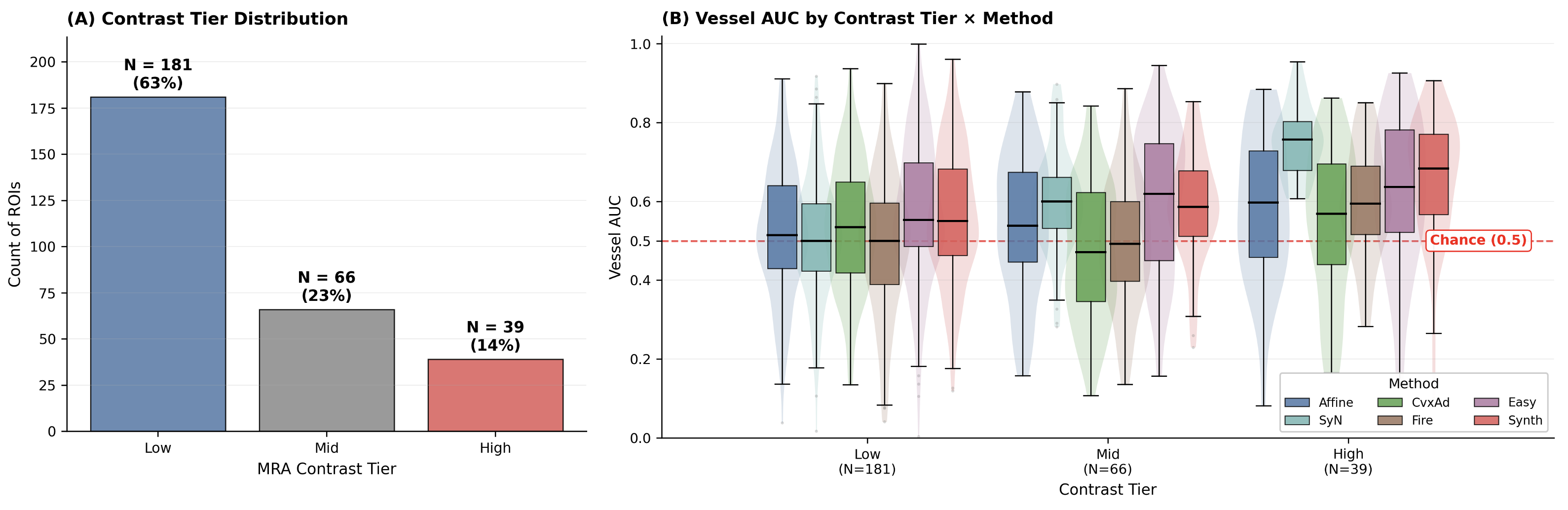}
    \caption{\textbf{Contrast-stratified analysis.} (A) Distribution of ROIs across Low, Mid, and High contrast tiers. (B) Vessel AUC by method within each tier. The dashed line indicates chance-level vessel-background separability.}
    \label{fig:contrast_stratified}
\end{figure}

We next asked how Vessel AUC behaves across MRA contrast regimes. Fig.~\ref{fig:contrast_stratified}A shows that 181 ROIs (63\%) were Low contrast, 66 (23\%) Mid contrast, and 39 (14\%) High contrast. Vessel AUC showed clear method separation in the Mid and High tiers; in the High tier, ANTs SyN achieved the highest median AUC (approximately 0.76). In the Low tier, Vessel AUC converged toward chance across methods, reflecting limited discriminative signal when local vessel-background contrast was weak. Because both the contrast score and Vessel AUC are derived from local MRA intensities sampled at clinician-labeled vessel and background voxels, this stratification should be interpreted as an assessment of when Vessel AUC is informative, not as a causal analysis of acquisition quality. Thus, method separation was most interpretable in the Mid and High tiers.

Importantly, the Mid tier was not simply a failure regime. Within the SyN-defined Mid tier ($N = 66$), ANTs SyN achieved a median Vessel AUC of 0.600 [IQR: 0.532-0.661], with 33/66 ROIs (50.0\%, patient-level bootstrap 95\% CI: 38.2-62.1\%) exceeding AUC $= 0.60$. However, measurable image-level vessel separability did not guarantee downstream segmentation-based localization. Among 65 Mid tier ROIs with valid SyN vessel predictions, 36/65 ROIs (55.4\%, 95\% Wilson CI: 43.3-66.8\%) missed the annotation under a 2-mm criterion ($d_{\mathrm{ann}\to\mathrm{Pred}} > 2$mm). Even among ROIs with $\mathrm{AUC}_{\mathrm{SyN}} > 0.60$, 16/33 (48.5\%) missed the annotation; conversely, 16/36 Mid-tier miss cases (44.4\%) still exceeded AUC $= 0.60$. Within the Mid tier, Vessel AUC and $d_{\mathrm{ann}\to\mathrm{Pred}}$ were not significantly associated (Spearman $\rho = -0.12$, $p = 0.35$). These results separate image-level vessel separability from post-registration model-level vessel extraction.

\subsection{FOV Mismatch Limits the Benefit of Deformable Refinement}
\begin{figure}[t]
    \centering
    \includegraphics[width=\columnwidth]{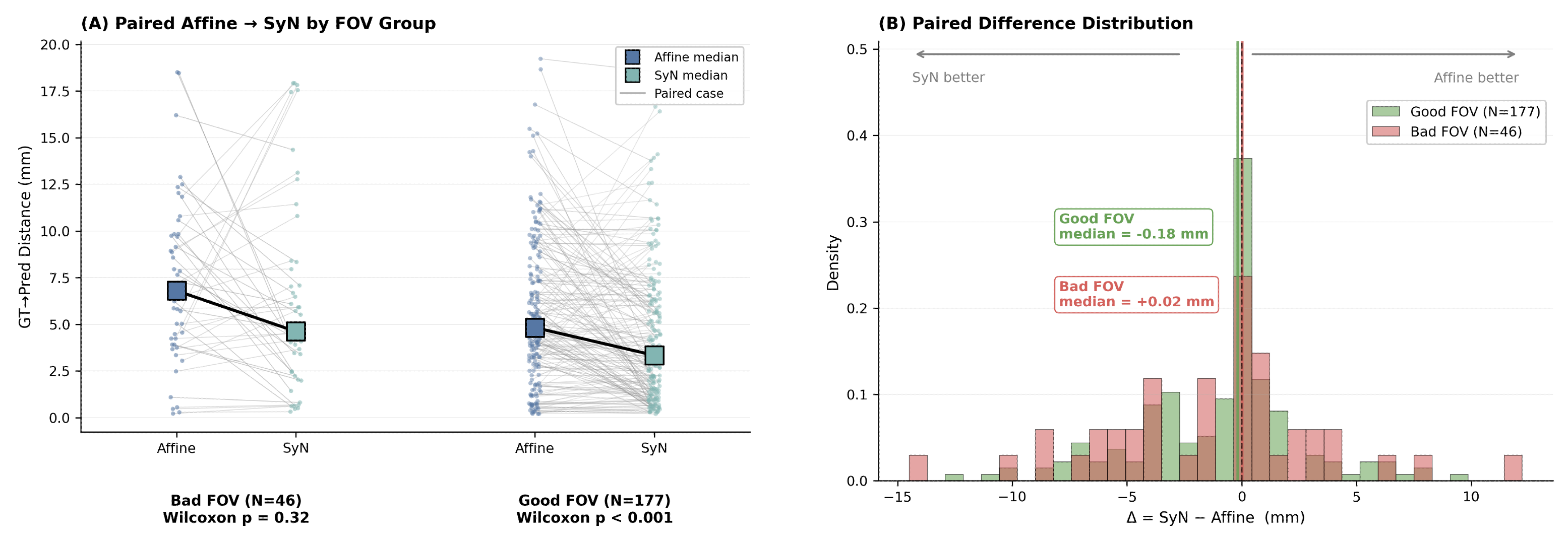}
    \caption{\textbf{Effect of FOV mismatch on deformable refinement.} (A) Paired Affine$\to$SyN trajectories; method medians are shown as filled squares. (B) Distribution of paired deltas ($\Delta=\mathrm{SyN}-\mathrm{Affine}$) by FOV group; vertical dashed lines mark group medians.}
    \label{fig:fov_mismatch}
\end{figure}

Finally, we examined whether deformable refinement remained beneficial under MRI–MRA FOV mismatch. The paired Affine-versus-SyN analysis included 223 ROIs: 177 Good-FOV and 46 Bad-FOV ROIs. In the Good-FOV subset, SyN produced a small but directionally consistent reduction in $d_{\mathrm{ann}\to\mathrm{Pred}}$ relative to affine alignment (median paired $\Delta=-0.18$mm, Wilcoxon $p<0.001$; patient-level bootstrap 95\% CI for the median: $[-0.55,+0.01]$mm). In the Bad-FOV subset, this benefit was not observed (median paired $\Delta=-0.07$mm, Wilcoxon $p=0.32$; bootstrap 95\% CI: $[-3.63,+0.65]$mm), and individual trajectories moved in both directions (Fig.~\ref{fig:fov_mismatch}).

These results indicate that deformable refinement provides at most a small and FOV-dependent improvement over affine alignment in this clinical setting. The wide Bad-FOV interval reflects limited sample size and high case-to-case variability, suggesting attenuation rather than reversal of the deformable-registration benefit. Robust affine initialization remains essential, and deformable outputs should be interpreted in light of shared anatomical coverage rather than assumed to improve local neurovascular alignment uniformly.

\section{Discussion}

This retrospective ROI-centered benchmark shows that TN MRI-MRA registration should be treated as a task-specific local neurovascular visualization problem rather than a generic whole-brain registration problem.

\textbf{Metric design under partial clinical annotations.} No single metric captures registration quality comprehensively. Vessel AUC and ROI NMI rank methods differently; $d_{\mathrm{ann} \to \mathrm{Pred}}$ and $d_{{\mathrm{Pred} \to \mathrm{ann}}}$ rank methods in opposite directions; and predicted vessel volume differs by more than an order of magnitude across methods. The volume-distance coupling analysis ($\rho = -0.53$ versus $\rho = +0.08$) makes clear that one-sided distance metrics can systematically favor methods with larger vessel predictions under partial clinical annotations. TN-focused benchmarks should therefore report Vessel AUC, ROI NMI, and symmetric vessel distance $d_{\mathrm{sym}}$ jointly with $|V_{\mathrm{Pred}}|$, rather than ranking methods on a single one-sided criterion.

\textbf{Image-level visibility versus model-level extraction.} The Mid-tier analysis shows a gap between local image conspicuity and downstream vessel extraction. These ROIs were not uniformly poor-quality: under ANTs SyN, the median Vessel AUC was 0.600. Nevertheless, downstream localization remained unreliable. Even among Mid-tier ROIs with $\mathrm{AUC}_{\mathrm{SyN}} > 0.60$, 48.5\% missed the clinician-annotated vessel segment under the 2-mm criterion, and Vessel AUC was not significantly associated with $d_{\mathrm{ann}\to\mathrm{Pred}}$ ($\rho = -0.12$, $p = 0.35$; Fig.~\ref{fig:case_example}B). Failures were driven by non-empty predictions that did not spatially match the annotated segment, rather than by empty masks. Thus, local vessel-background separability on the registered MRA is not interchangeable with successful downstream vessel extraction. Because vessel segmentation is applied after registration, segmentation-derived distances reflect the combined registration–segmentation pipeline rather than registration error alone. A future segment-once-then-warp analysis, in which vessels are extracted from the original MRA and then warped by each registration field, could further separate registration geometry from segmentation-model sensitivity.

\textbf{Contrast-stratified evaluation as a methodological necessity.} Vessel AUC measures local vessel-background separability in the warped MRA (Eq.~\eqref{eq:auc}) and cannot distinguish methods when local contrast is intrinsically weak. In this cohort, the Low-contrast tier comprised 63\% of ROIs and drove Vessel AUC toward chance for all methods, whereas clear method separation appeared in the Mid and High tiers. Thus, contrast stratification should not be interpreted causally as contrast being more important than method choice; rather, it identifies when Vessel AUC is informative. Reporting Vessel AUC without contrast stratification can mask genuine method differences in clinical cohorts dominated by low-contrast ROIs.

\begin{table}[t]
\centering
\caption{Qualitative method profile across the evaluated registration methods. Numerical values are reported in Table~\ref{tab:overall_metrics}.}
\label{tab:method_profile}
\small
\setlength{\tabcolsep}{5pt}
\renewcommand{\arraystretch}{1.0}
\begin{tabularx}{\columnwidth}{@{}l
>{\raggedright\arraybackslash}X
>{\raggedright\arraybackslash}X@{}}
\toprule
\textbf{Method} & \textbf{Observed strength} & \textbf{Main trade-off} \\
\midrule
\textbf{ANTs SyN}
& Most balanced across all metrics
& Higher CPU runtime; batch-parallelizable \\
ConvexAdam
& Lowest symmetric distance $d_{\mathrm{sym}}$
& Requires affine pre-alignment \\
SynthMorph
& Best one-sided $d_{\mathrm{ann}\to\mathrm{Pred}}$; high Vessel AUC
& Large prediction volume inflates one-sided coverage \\
EasyReg
& High Vessel AUC; fast learning-based inference
& Large, variable prediction volume \\
FireANTs
& Highest ROI NMI (local intensity match)
& Highest reverse distance $d_{\mathrm{Pred}\to\mathrm{ann}}$ \\
ANTs Affine
& Reliable baseline under FOV mismatch
& No deformable refinement \\
\bottomrule
\end{tabularx}
\begin{minipage}{\columnwidth}
\scriptsize
\vspace{4pt}
\textit{Segmentation-derived distances are downstream metrics, not direct registration error.}
\end{minipage}
\end{table}

\begin{figure}[t]
    \centering
    \includegraphics[width=\columnwidth]{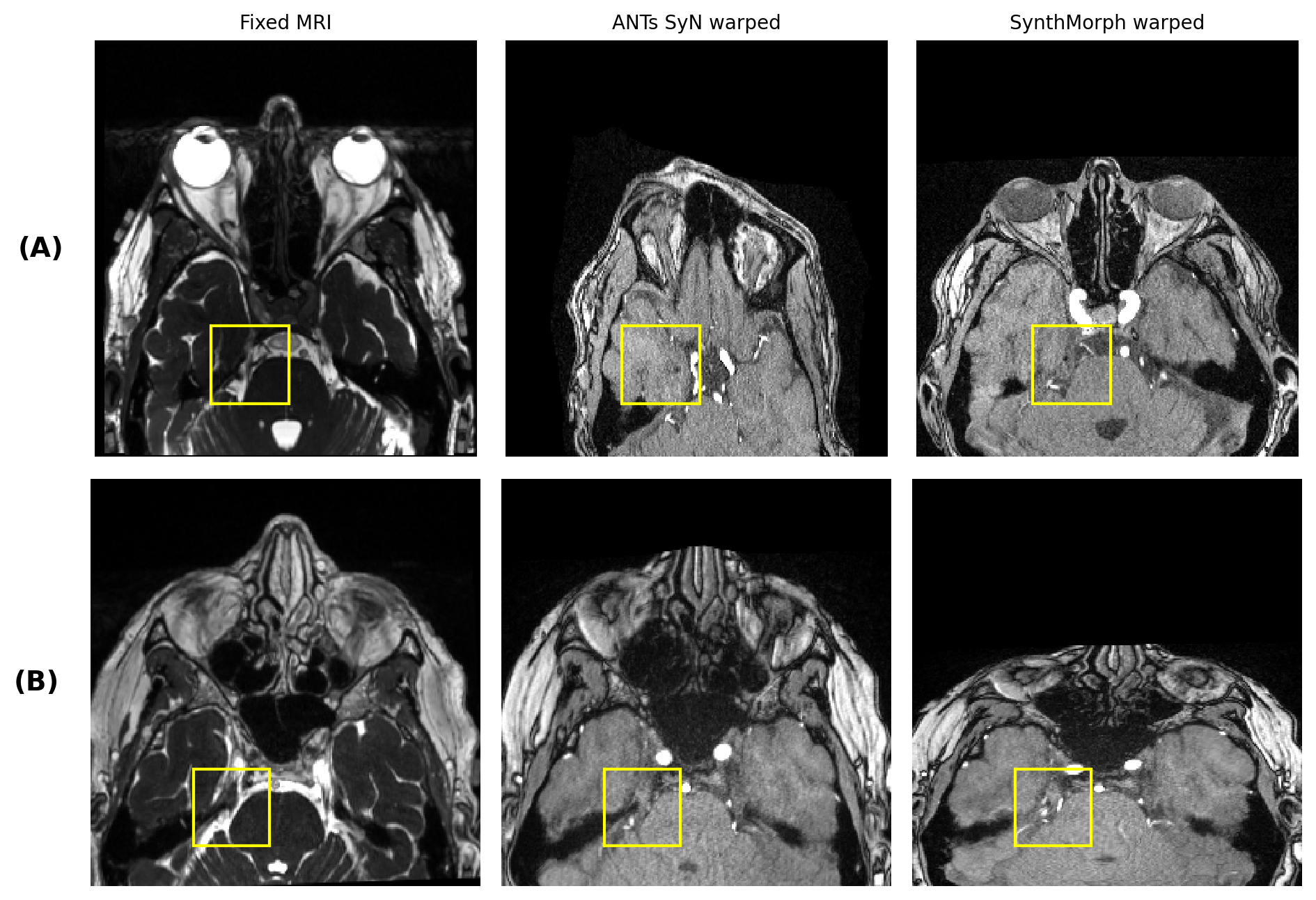}
    \caption{Whole-brain registration failures can dissociate from one-sided ROI distance metrics. Each row shows the fixed structural MRI, ANTs-SyN warped MRA, and SynthMorph warped MRA at the trigeminal ROI. \textbf{(A)} Gross misalignment rated non-evaluable. \textbf{(B)} SynthMorph global distortion with preserved local ROI proximity, while ANTs SyN remains anatomically plausible. Since $d_{\mathrm{ann}\to\mathrm{Pred}}$ is computed only within the ROI, globally failed but locally proximate predictions can yield deceptively low one-sided distances.}
    \label{fig:failure_modes}
\end{figure}

\textbf{Robust affine alignment is a prerequisite.} FOV-stratified paired comparisons showed that deformable refinement provided a statistically significant but small benefit over affine alignment in Good-FOV ROIs ($p < 0.001$), while this benefit was attenuated in Bad-FOV ROIs ($p = 0.32$). The Good-FOV bootstrap CI nearly touched zero, indicating substantial ROI-level heterogeneity and modest average improvement. In clinical data with partial modality overlap, robust affine initialization should therefore be treated as a prerequisite, and methods that assume strong pre-alignment, such as ConvexAdam, require an explicit affine stage.

\textbf{Blinded reader study and the venous-offender hypothesis.} The Low-contrast tier may reflect limited TOF-MRA visibility, including possible enrichment for venous offenders since TOF-MRA preferentially depicts fast arterial flow. In the blinded reader study of 100 stratified ROIs, 66 were clinically evaluable and 34 were not; among evaluable ROIs, readers identified 42 arterial, 9 venous, 14 mixed arterial-venous, and 1 with no definite vessel. Thus, 23/66 evaluable ROIs (34.8\%) had some venous component, supporting the hypothesis that part of the Low-contrast regime may reflect venous-offender biology rather than generic acquisition failure. However, TOF-MRA does not reliably depict venous anatomy, and this reader-based vessel typing is hypothesis-generating rather than a substitute for complementary venous imaging or intraoperative confirmation.

\textbf{Clinical evaluability exposes a metric confound.} Reader review also showed that favorable one-sided ROI distances can coexist with clinically unusable whole-brain registration. Across 100 reviewed ROIs, SynthMorph produced globally shifted or failed registrations in 71 cases; among the 66 ANTs-SyN-evaluable ROIs, 48 still showed SynthMorph whole-brain misalignment. As shown in Fig.~\ref{fig:failure_modes}, predicted vessels could fall close to the annotation within the small trigeminal ROI despite global distortion, yielding deceptively low $d_{\mathrm{ann}\to\mathrm{Pred}}$ values. This explains why SynthMorph had the lowest median $d_{\mathrm{ann}\to\mathrm{Pred}}$ in Table~\ref{tab:overall_metrics} despite frequent visual registration failure, and reinforces that segmentation-derived one-sided distances must be interpreted jointly with whole-brain registration quality.

\textbf{Balanced performance of ANTs SyN.} Across the six methods, ANTs SyN showed the most balanced profile (Table~\ref{tab:method_profile}). It was not the lowest performer on any metric, achieved the highest median Vessel AUC in the High-contrast tier ($\sim$0.76), and produced moderate predicted vessel volumes ($|V_{\mathrm{Pred}}|$ median 1493). In contrast, SynthMorph and EasyReg achieved lower $d_{\mathrm{ann}\to\mathrm{Pred}}$ and higher Vessel AUC but produced much larger predicted vessel trees, inflating $d_{\mathrm{Pred}\to\mathrm{ann}}$. ConvexAdam had the lowest $d_{\mathrm{sym}}$ but required external affine pre-alignment, while FireANTs showed high $d_{\mathrm{Pred}\to\mathrm{ann}}$ despite favorable $d_{\mathrm{ann}\to\mathrm{Pred}}$, again reflecting prediction-extent effects. ANTs SyN was slower, requiring approximately 20 minutes per MRI–MRA pair, but cases were independent and batch-parallelizable. These results support ANTs SyN as a reasonable baseline for local overlay generation when contrast and FOV coverage are adequate, while faster or more detection-sensitive alternatives may be preferable for other task-specific priorities.

\textbf{Clinical implications.} MRI–MRA co-registration may improve TN surgical evaluation by integrating structural nerve detail with vascular trajectories near the trigeminal nerve, a relationship directly relevant to MVD planning and prognosis. Prior work has demonstrated the clinical value of fused high-resolution MRI, TOF-MRA, and 3D visualization for posterior-fossa neurovascular assessment~\cite{Miller2008PreoperativeVisualization,Docampo2015NeurovascularStudy3T,Granata2013VirtualMRI,Dolati2015PreoperativeSegmentation,Yao2018VirtualRealityMVD,Gamaleldin2020FusedTOFCISS,Pham2021FusedSpaceTOFMRA,Hastreiter2022DataFusion3D,Huang2024MRVE}. However, the registration accuracy needed for local trigeminal ROI interpretation has not been systematically quantified.

Our ROI-centered benchmark addresses this gap by evaluating registration behavior directly at the trigeminal nerve in a 149-patient MVD cohort, using local image-based and vessel-localization metrics. When local MRA contrast is adequate and MRI–MRA FOV coverage is comparable, ANTs SyN provides a reasonable offline baseline for local overlay generation. MRA integration may also strengthen MRI-only neurovascular segmentation by adding vascular conspicuity to structural nerve information, potentially improving quantitative assessment of NVC morphology and prognosis~\cite{HalbertElliott2025Segmentation,Wang2026NVCQuantification}.

Because TOF-MRA is more sensitive to arterial than venous flow, MRI–MRA fusion may also help generate hypotheses about compressive vessel identity. Vessels visible on both structural MRI and TOF-MRA are more suggestive of arterial flow, whereas neurovascular contact visible on MRI but faint or absent on TOF-MRA may raise suspicion for low-flow venous structures. This distinction is clinically relevant because venous compression is associated with less favorable MVD outcomes than arterial compression~\cite{Nair2023ArteryVein}. However, TOF-MRA does not reliably depict venous anatomy, so vessel-type interpretation from this pipeline remains non-definitive without complementary venous imaging or intraoperative confirmation.

Learning-based methods such as SynthMorph and EasyReg remain useful when speed, throughput, or vessel-detection sensitivity is prioritized, but their larger predicted vessel volumes require joint interpretation with extent-sensitive distance metrics. In cases with poor local contrast, the current registration–segmentation pipeline is unlikely to provide reliable vessel localization without improved acquisition or contrast-aware extraction. In cases with substantial FOV mismatch, deformable refinement should not be assumed to improve local alignment; affine alignment should remain an essential reference, and deformable outputs should undergo local quality control.

\subsection{Limitations}

This study has several limitations. It is retrospective and single-center; although it includes multiple scanners, vendors, and field strengths, it remains predominantly Siemens 3T, and external validation on more heterogeneous TN cohorts would further assess generalizability. The annotations contain clinically relevant partial vessel segments rather than exhaustive vessel masks, matching the surgical-review workflow but complicating geometric evaluation because predicted vessel trees may extend beyond the labeled segment; one-sided distances should therefore be interpreted jointly with predicted vessel volume. Vessel segmentation was performed after registration, so the resulting masks may be affected by interpolation, deformation, resampling, and method-specific warped-image appearance; segmentation-derived distances therefore reflect the registration–segmentation pipeline rather than registration error alone. The statistical analyses are exploratory and do not include mixed-effects modeling or correction for multiple stratified analyses, and annotation and reader-review variability were not fully quantified because both used consensus workflows. Future work should incorporate intraoperative offending-vessel confirmation, formal NVC grading, MVD outcome modeling, and external validation to better relate ROI-centered MRI–MRA registration to operative findings and patient outcomes.

\section{Conclusion}

We presented an ROI-centered benchmark for MRI–MRA registration in preoperative TN neurovascular visualization. In 149 patients and 298 clinician-reviewed ROIs, we evaluated six registration pipelines using local image-based metrics and segmentation-derived vessel-localization measures. Our findings show that TN MRI–MRA fusion cannot be judged by generic whole-brain registration criteria: image-level vessel separability and downstream vessel localization are non-interchangeable, one-sided vessel distances are confounded by predicted vessel extent under partial annotations, Vessel AUC is informative mainly when local MRA contrast is sufficient, and deformable refinement provides only a small, FOV-dependent benefit over affine alignment. These results support evaluating TN MRI–MRA registration as a local, vessel-aware, contrast-sensitive, and FOV-aware visualization task. Joint reporting of image-based and segmentation-derived metrics, prediction volume, contrast regime, FOV compatibility, and registration quality control provides a practical foundation for future ROI-aware neurovascular visualization methods.


\bibliographystyle{unsrtnat}
\bibliography{refs}

\clearpage
\setcounter{subsection}{0}
\renewcommand{\thesubsection}{S\arabic{subsection}}
\renewcommand{\theHsubsection}{S\arabic{subsection}}
\setcounter{table}{0}
\renewcommand{\thetable}{S\arabic{table}}
\renewcommand{\theHtable}{S\arabic{table}}
\setcounter{figure}{0}
\renewcommand{\thefigure}{S\arabic{figure}}
\renewcommand{\theHfigure}{S\arabic{figure}}
\section*{Supplementary Material}

This Supplementary Material provides additional methodological details for the MRI--MRA registration benchmark described in the main manuscript. Sections S1--S3 describe cohort selection, imaging acquisition, and ROI annotation; Sections S4--S5 describe registration settings, computing resources, and VesselFM inference; and Sections S6--S7 describe contrast stratification, descriptive thresholds, and statistical analysis. Supplementary Table~\ref{tab:supp_s1} summarizes the imaging acquisition parameters. References to numbered sections and figures refer to the main manuscript unless otherwise indicated.

\subsection{Cohort Selection and Demographics}
\label{supp:cohort}
\textbf{Eligibility.} We retrospectively studied a single-institution TN cohort consisting of patients who underwent microvascular decompression (MVD) between January 2020 and December 2022; diagnostic imaging dates ranged from October 2019 to December 2022. Inclusion required paired structural MRI and TOF-MRA from the same preoperative diagnostic imaging encounter and clinician-provided trigeminal annotations.

\textbf{Cohort flow.} From an initial 159 paired MRI-MRA scans, 6 cases were excluded during preprocessing because of DICOM read errors, autocropping artifacts, or related technical failures, yielding 153 successfully preprocessed pairs. A further 4 cases were excluded because trigeminal annotations existed but the corresponding MRA volume was missing or could not be reliably paired, resulting in a final analysis cohort of 149 patients (mean age $56.6 \pm 14.0$ years, range 20.1-82.4; 85 female / 64 male; 88 right-sided / 61 left-sided symptomatic presentation, no bilateral; 7 patients [4.7\%] with multiple sclerosis as a secondary cause of TN).

\textbf{Analysis cohort.} All final-cohort patients subsequently underwent MVD, and all imaging preceded surgery. All patients had bilateral ROIs annotated, yielding 298 trigeminal ROIs in total.

\textbf{Ethics approval.} The study was approved by the Johns Hopkins Medicine Institutional Review Board (IRB00338945). Informed patient consent was waived given the retrospective nature of the study.

\subsection{Imaging Acquisition Parameters}
\label{supp:acquisition}
\textbf{Sequences and scanners.} Structural MRI (CISS-type) was used to visualize the trigeminal nerve and surrounding soft tissue, and TOF-MRA was used to depict vascular structures. In brief, 98.0\% of scans were acquired on Siemens scanners (predominantly Skyra) and the remainder on GE Healthcare; 92.6\% of MRI and 91.9\% of MRA scans were acquired at 3T, with the remainder at 1.5\,T. Structural MRI used a 3D CISS-type sequence (median TR/TE 5.45/2.42\,ms; median in-plane spacing 0.597mm; median slice thickness 0.60mm), and TOF-MRA used a 3D multi-slab TOF sequence (median TR/TE 22.00/3.78\,ms; median in-plane spacing 0.260mm; median slice thickness 0.50mm).

\textbf{Acquisition timing and reconstructions.} In the final analysis cohort, MRI acquisition dates ranged from January 2020 to December 2022 and MRA acquisition dates from October 2019 to December 2022. A small subset of structural MRI scans had DICOM slice thickness greater than 0.6mm, reflecting non-CISS reconstructions; however, ROI annotation and all analyses were performed on the native high-resolution CISS series. Detailed per-modality parameters are reported in Supplementary Table~\ref{tab:supp_s1}.

\begin{table*}[!t]
\centering
\caption{Imaging acquisition parameters per modality (N = 149 patients). Continuous parameters are summarized as median [Q1, Q3] (range). Categorical parameters are summarized as count (\%). All parameters are derived from DICOM headers; in-plane and slice spacing are confirmed from NIfTI volume headers used for registration.}
\label{tab:supp_s1}
\small
\begin{tabular}{p{4.6cm} p{5.0cm} p{5.0cm}}
\toprule
\textbf{Parameter} & \textbf{Structural MRI (CISS-type)} & \textbf{TOF-MRA} \\
\midrule
\multicolumn{3}{l}{\textit{Vendor and scanner}} \\
\midrule
Manufacturer (Siemens / GE)            & 146 (98.0\%) / 3 (2.0\%)              & 146 (98.0\%) / 3 (2.0\%) \\
Most common model                      & Skyra: 115 (77.2\%); Verio: 12 (8.1\%); Aera: 9 (6.0\%); MAGNETOM Vida: 5 (3.4\%); other: 8 (5.4\%)
                                       & Skyra: 112 (75.2\%); Verio: 13 (8.7\%); Aera: 9 (6.0\%); MAGNETOM Vida: 7 (4.7\%); other: 8 (5.4\%) \\
Field strength (3.0\,T / 1.5\,T)       & 138 (92.6\%) / 11 (7.4\%)             & 137 (91.9\%) / 12 (8.1\%) \\
\midrule
\multicolumn{3}{l}{\textit{Sequence parameters (DICOM)}} \\
\midrule
Predominant series description         & POST/PRE CISS SAG MPR (and reconstructions thereof) & TOF\_3D\_multi-slab (130/149); MRA cow (7); other configurations (12) \\
Repetition time TR (ms)                & 5.45 [5.43, 5.46] (5.00-7.84)        & 22.00 [22.00, 22.00] (19.00-25.00) \\
Echo time TE (ms)                      & 2.42 [2.41, 2.43] (2.04-3.66)        & 3.78 [3.78, 3.78] (3.40-7.15) \\
Slice thickness, DICOM (mm)            & 0.60 [0.60, 0.60] (0.59-1.00)        & 0.50 [0.50, 0.60] (0.40-1.20) \\
\midrule
\multicolumn{3}{l}{\textit{Volume geometry (NIfTI)}} \\
\midrule
In-plane spacing (mm)                  & 0.597 [0.597, 0.597] (0.372-0.880)   & 0.260 [0.260, 0.288] (0.260-0.482) \\
Slice spacing (mm)                     & 0.594 [0.594, 0.594] (0.150-0.859)   & 0.500 [0.500, 0.600] (0.400-0.800) \\
Median matrix shape                    & $254 \times 256 \times 164$           & $646 \times 768 \times 165$ \\
\midrule
\multicolumn{3}{l}{\textit{Acquisition timing}} \\
\midrule
Study date range                       & 2020-01-05 to 2022-12-20              & 2019-10-08 to 2022-12-20 \\
\bottomrule
\end{tabular}
\end{table*}

\subsection{ROI Annotation Protocol}
\label{supp:annotations}
\textbf{Annotation workflow.} ROI centroids were identified using ITK-SNAP, and per-voxel labels were drawn manually on cropped structural MRI volumes using the Napari Python toolkit. The final annotations used for this benchmark were defined on $48 \times 48 \times 48$ ROI crops centered on the cisternal trigeminal nerve, which is the unit of analysis used throughout the study. Each volume received a first-pass segmentation followed by independent second-pass review and editing before finalization. Annotations were drawn on structural MRI alone, using vessel-related signal voids and local neurovascular anatomy visible on MRI, without overlay on MRA or any registration output. The annotation protocol was therefore independent of the registration methods evaluated here. The labels are intended to mark clinically relevant candidate vessel segments rather than to provide an exhaustive vascular tree; this partial-annotation property is central to the metric design described in Section~\ref{sec:eval-metrics}. The two-pass workflow produced consensus rather than parallel annotations, which limits direct quantification of inter-annotator variability.

\textbf{Operational definition of the clinically relevant vessel segment.} Vessel labels were drawn to capture all vessels in contact with or in close proximity to the trigeminal nerve, with an emphasis on the contact region. Distinctly visible non-contacting vessels reasonably close to the nerve were also segmented; very peripheral or low-signal vessels (e.g., short tracks visible only across 3-4 contiguous slices) were not labeled. Because the protocol focuses on the contact region, vessel coverage may be incomplete near the periphery of the $48 \times 48 \times 48$ ROI for some cases. This is the operational source of the partial-annotation property described next.

\textbf{Implications for vessel-proximity metrics.} An important property of this protocol is that vessel labels mark only the vessel segment considered clinically relevant to the trigeminal nerve within the ROI, rather than every vessel voxel that could in principle be labeled on MRA. In contrast, VesselFM and other foundation vessel segmentation models produce a full vascular trace that typically extends beyond the annotated segment. This mismatch is intrinsic to the clinical annotation workflow and is a critical caveat for all vessel-proximity metrics reported in the main text; we revisit it quantitatively in Section~\ref{sec:volume-coupling}.

\subsection{Registration Settings and Computing Resources}
\label{supp:reg-config}
\textbf{Registration settings.} ANTs Affine and SyN used a multi-resolution schedule with Mattes mutual information for rigid/affine stages and cross-correlation for SyN refinement. ConvexAdam used the \texttt{ConvexAdam\_MIND\_brain\_default} configuration with MIND-SSC descriptors and 99.5th-percentile intensity normalization. FireANTs was run with moments-based center-of-mass/rigid alignment followed by multi-resolution greedy deformable registration. EasyReg and SynthMorph were run with FreeSurfer default configurations. No external brain or registration masks were supplied to any method.

\textbf{Software and hardware.} All experiments were run under Ubuntu 24.04 with Python 3.10. The registration methods used the following versions: ANTs 2.6.2, FireANTs 1.0.0, ConvexAdam (MIND-SSC implementation, accessed through the publicly released Medical Image Registration (MIR) toolbox~\cite{Chen2024MIR}), and EasyReg/SynthMorph from FreeSurfer 8.1.0. Learning-based and GPU-accelerated methods used PyTorch 2.10.0 with CUDA 12.8. The compute server (ldr01 at our institution) had 2 AMD EPYC 9354 CPUs (128 logical threads), 1.5\,TB RAM, and 8 NVIDIA RTX PRO 6000 Blackwell Max-Q GPUs (96\,GB each, driver 580.126). Cases are independent and were processed in parallel across CPU cores or GPUs as appropriate; the effective per-case wall-clock time depended on the batch parallelization rather than on the single-case algorithmic cost. ANTs SyN was the runtime-dominant step, on the order of $\sim$20\,min CPU per MRI-MRA pair in our deployment, but offline batch processing across the 149-patient cohort completed within a small number of days.

\textbf{Memory-related fallback.} All six methods produced warped MRA outputs for all 149 patients (298 ROIs). Four patients had high-resolution MRI reconstructions with in-plane voxel spacing 0.15-0.17mm (versus the cohort-typical 0.59mm CISS), yielding preprocessed volumes that exceeded single-GPU memory for ConvexAdam; these cases were processed using an identical CPU-based ConvexAdam pipeline and completed successfully.

\subsection{VesselFM Inference and Postprocessing}
\label{supp:vesselfm}
\textbf{Model and application.} We applied VesselFM~\cite{Wittmann2025VesselFM} (v1.0, \url{https://github.com/bwittmann/vesselFM}) to each warped MRA volume in MRI space to produce a whole-brain vessel segmentation, which was then cropped to the matched trigeminal ROI and compared against the clinician annotation. We used the publicly released model (\texttt{dyn\_unet\_base}) without retraining or fine-tuning, and applied the same inference configuration to every method's warped MRA.

\textbf{Inference and postprocessing.} Each warped MRA was preprocessed by 1st-99th-percentile intensity normalization to $[0, 1]$ with clipping, then run through sliding-window inference with a $128^3$ patch size, 0.5 patch overlap, and constant-mode merging. No test-time augmentation was used. The predicted probability map was binarized at threshold 0.5, after which the repository's default postprocessing removed connected components smaller than 500 voxels (face-edge-vertex connectivity); no additional filtering was applied.

\textbf{ROI-based evaluation.} The resulting whole-brain vessel mask was resampled into the same trigeminal $48^3$ ROI sampling grid used for image-based metrics. Vessel-proximity metrics required at least one predicted vessel voxel in the ROI for both compared methods; ROI-method observations with empty predicted vessel mask were excluded only from those metrics, while contrast and Vessel AUC, which depend only on the warped MRA intensity, were retained.

\subsection{Contrast Stratification and Exclusions}
\label{supp:contrast}
\textbf{Cross-method consistency.} Across the six registration methods, per-ROI contrast values shared a similar overall scale (pairwise Pearson 0.41-0.74 on the 278 ROIs with valid contrast measurements under all six methods) but the per-ROI rank ordering of contrast was only weakly preserved between methods (pairwise Spearman 0.11-0.51, median 0.24).

\textbf{Reference definition and exclusions.} Using the SyN contrast as the canonical reference therefore reflects an analysis choice rather than a method-invariant ROI property, and the per-method Vessel AUC distributions in Section~\ref{sec:contrast-strat} are reported within the same SyN-defined tiers to ensure comparability. Of 298 total ROIs, 286 yielded valid contrast measurements; 12 ROIs were excluded because the ANTs SyN deformable warp placed the TN-centered sampling grid outside the warped MRA coverage.

\subsection{Descriptive Thresholds and Statistical Analysis}
\label{supp:stats}
\textbf{Descriptive thresholds.} For cross-tabulation of image-level versus downstream outcomes within the Mid-contrast tier (Section~\ref{sec:contrast-strat}), we additionally classified ROIs by two descriptive thresholds. An ROI was classified as image-level high-AUC if its Vessel AUC under ANTs SyN exceeded $0.60$, chosen as a pragmatic cutoff above the $0.5$ chance level rather than an optimized decision threshold. A non-empty post-registration vessel prediction was classified as a downstream miss if $d_{\mathrm{ann}\to\mathrm{Pred}} > 2$mm; this threshold corresponds to roughly four voxels on the $0.47$mm ROI grid and was used solely for descriptive cross-tabulation, not for primary method ranking.

\textbf{FOV stratification threshold.} The 20th-percentile FOV cutoff ($r_{\min} \le 0.76$) was used to define an interpretable severe-mismatch subgroup; the primary conclusion is the attenuation pattern of deformable benefit rather than the specific numerical cutoff, and this threshold-based stratification should be interpreted as descriptive.

\textbf{Statistical analysis.} Distributions are summarized as medians with interquartile ranges. Volume-distance coupling was assessed using Spearman rank correlation. Paired Affine-versus-SyN comparisons were assessed using the Wilcoxon signed-rank test. Because each patient contributes bilateral ROIs that are not independent (same scanner, same anatomy), we report patient-level bootstrap 95\% confidence intervals (1000 resamples) for all key descriptive and inferential statistics. Each bootstrap iteration resampled the 149 patients with replacement; resampled patients always contributed both ipsilateral and contralateral ROIs together, naturally preserving the within-patient correlation. Because the analyses are exploratory and the same data are used for multiple stratifications, we do not adjust for multiple comparisons; reported $p$-values and CIs should be interpreted with this in mind.


\end{document}